\documentclass[11pt]{article}
\usepackage{template_style}
\usepackage{colortbl}   
\usepackage{multirow}
\usepackage{adjustbox}  
\usepackage[capitalise,noabbrev]{cleveref}
\definecolor{oursred}{HTML}{C8102E}

\begin{document}
\ogsettitle{OralDetect: An Open and Unified Model Family for \\ Multi-Modality Dental Object Detection}

\ogsetauthors{Siyuan Dai$^{1*}$, Jing Hao$^{2*\dagger}$, Jiahao Bao$^{2}$, Xinpeng Lu$^{3}$, Yuxuan Fan$^{4}$, Liang You$^{1}$,
              Shuaibo Li$^{5}$, Jie Huang$^{6}$, Ming Hu$^{7}$, Kuo Feng Hung$^{2}$, Liang Zhan$^{1}$\textsuperscript{\,\scriptsize\faEnvelope[regular]}}

\ogsetaffiliations{$^{1}$University of Pittsburgh,
                   $^{2}$The University of Hong Kong,
                   $^{3}$University of Science and Technology of China,
                   $^{4}$Nanyang Technological University,
                   $^{5}$The Chinese University of Hong Kong,
                   $^{6}$Nankai University,
                   $^{7}$Shanghai Artificial Intelligence Laboratory\\[3pt]
                   {\small $^{*}$Equal contribution \quad $^{\dagger}$Project leader \quad \textsuperscript{\,\scriptsize\faEnvelope[regular]}\,Corresponding author}}

\ogsetteam{OralGPT-Family}

\ogsetemail{isjinghao@gmail.com}

\ogsetlinkA{https://github.com/isjinghao/OralGPT}
\ogsetlinkB{https://huggingface.co/OralGPT}
\ogsetlinkC{https://github.com/isjinghao/OralGPT}

\ogsetdate{2026-11-23}
\ogsetreporttype{Technical Report}
\ogsetstatus{OralDetect technical report}

\ogsetabstract{%
  Vision-language foundation models excel on general-domain, vision-centric benchmarks, yet computer-aided oral image analysis is clinically important and practical but remains underexplored. General models lack oral domain knowledge, and oral diagnosis relies on multiple fine-grained semantics. Especially in the cross-modality clinical setting, where the same textual concept maps to different visual evidence across modalities, e.g., ``caries'' denotes a discolored surface in an intraoral photograph but a radiolucency on a radiograph. Consequently, general-domain detectors transfer poorly, and naively fine-tuning general foundation models with multi-modal datasets may conflate text semantics with visual semantics. To this end, we present OralDetect Suite, an open model family for open-vocabulary oral condition detection across modalities. We curate and purify 718,839 dental image-text pairs, 466,951 of them diagnostic images from eight modalities, and pretrain OralDetect, which couples an OralCLIP vision-language model with an OralBERT text encoder. To address the cross-modality gap, we inject a modality special token into the vision tower and a modality-conditioned semantic calibration module into the text tower. We further build cross-modality benchmarks for open-vocabulary detection and vision-language retrieval. Our proposed OralDetect family surpasses general-domain and domain-specific baselines on both benchmarks. All data, benchmarks, and models will be released to the community.}

\ogmakefrontmatter


\section{Introduction}
\label{sec:intro}

\begin{figure}[!htbp]
\centering
\includegraphics[width=\linewidth]{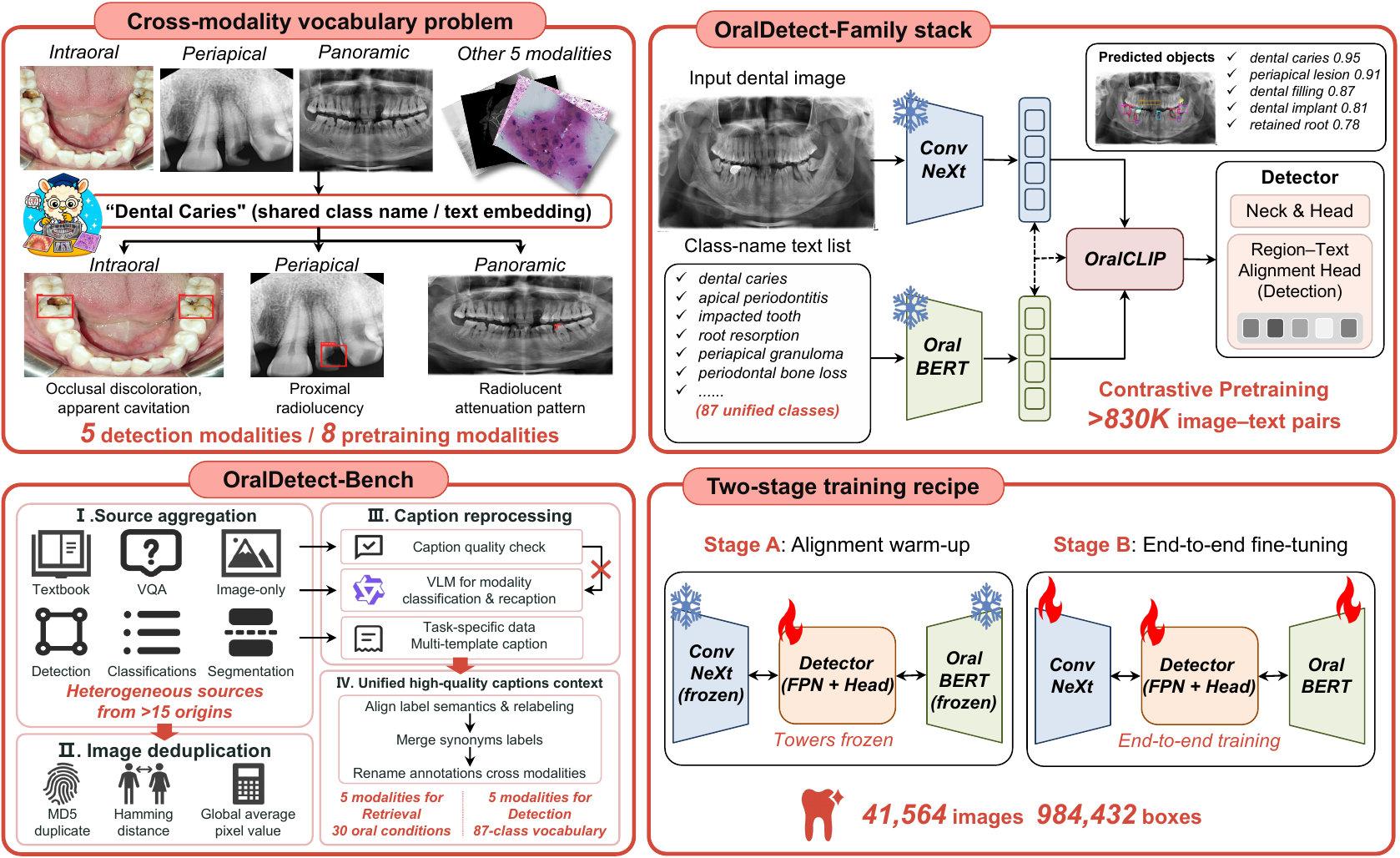}
\caption{Overview of the OralDetect Suite: a ConvNeXt-based vision tower and an OralBERT text tower are aligned by OralCLIP contrastive pretraining and transferred into a cross-modality open-vocabulary detector that scores each region against the $87$ class-name embeddings produced by the OralBERT text tower.}
\label{fig:main}
\end{figure}

Vision-language foundation models have advanced quickly on general-domain, vision-centric benchmarks \cite{clip,glip,yoloworld}, and an open-vocabulary detection model can detect a class named in free text without ever having been labelled at training time. However, such a mechanism in computer-aided oral image analysis is still underexplored, while it is clinically important and highly demanded. Clinicians and pathologists read across several imaging modalities, i.e., intraoral photography, panoramic X-ray, periapical radiography, cytology, and histopathology, and draw on expert oral knowledge with hundreds of conditions, tooth positions, quadrants, etc. Therefore, a suite of foundation models is needed in assisting clinical practice for open-vocabulary detection across multiple modalities. Meanwhile, general-domain foundation models are not a substitute, since their vision-language spaces \cite{clip,biomedclip} have barely seen dental terminology, so a general open-vocabulary detector can hardly read the semantics of an oral image precisely.

Moreover, the missing domain knowledge in the oral domain is a structural obstacle that differs from other professional domains. Oral diagnosis is inherently cross-modality, since one text concept is realised differently in different image modalities. The word ``caries'' picks out a property of a visible surface in an intraoral photograph, but a shortfall of mineral registered as attenuation in a panoramic radiograph, so the same text semantic has no single visual appearance, as shown in \cref{fig:main}~(a). In an open-vocabulary detector, the text embedding is the classifier, so a single embedding of ``caries'' trained across modalities is pulled toward a compromise between different visual semantics and fits none of them well. Naively increasing the data pool only sharpens the compromise. Ideally, the image-text space and the detector should be built on the same semantic space and explicitly calibrate the modality-dependent semantic gap between image and text.

To address this problem, we propose OralDetect Suite, an open family of dental foundation models built as a single stack. At its base are a dental vision tower and a dental text tower, OralBERT, pretrained on a corpus of influential dental textbooks. The two towers are aligned by contrastive image-text pretraining on ${>}710$K dental image-text pairs over eight modalities into OralCLIP, and transferred into a calibrated, retrieval-based OralDetect with a light detection neck and head, which is an open-vocabulary detector. Specifically, we introduce two modality-aware components to bridge the cross-modality gap: a modality-specific token in the vision tower and a modality-conditioned semantic calibration in the text tower. Notably, the retrieval-based two-tower design yields a fast detector compared with the DINO family, which benefits daily clinical use.

Furthermore, no multi-modality oral detection benchmark has been released. To evaluate this critical task comprehensively, we organise OralDetect-Bench, the first cross-modality oral-vision benchmark (\cref{fig:bench}). It contains two sub-benches, one for retrieval with 3,300 samples across 5 modalities, and another for detection with 3,076 images across 5 modalities. The detection bench has both a closed-set test under an 87-class vocabulary and an open-set test on two held-out intraoral datasets, dental plaque and dental crack. Against twelve detection baselines spanning closed-set and general open-vocabulary detectors, and five retrieval-centric baselines, our OralDetect family demonstrates superiority (\cref{fig:qual}). The main contributions are:

\begin{enumerate}
\item We propose an open family of dental foundation models, released as one stack: a dental vision tower and text tower, OralBERT, aligned into OralCLIP on 710K dental image-text pairs across eight modalities, and transferred into OralDetect, an open-vocabulary detector.
\item Our retrieval-based OralDetect is designed practically with two towers and a lightweight modality calibration module, yielding a fast and modality-sensitive detector.
\item We propose OralDetect-Bench with two sub-benches for CLIP-based and detection-based tasks, covering both closed-set and open-set evaluation. Our OralDetect family reports leading results against baselines.
\end{enumerate}

\section{Related Work}
\label{sec:related}

\textbf{Domain vision-language models.} Contrastive image-text pretraining \cite{clip} yields a joint space in which a class is named by a learnable embedding rather than a fixed index, and this space is what open-vocabulary detection consumes. Biomedical variants re-fit it to clinical language: MedCLIP \cite{medclip} decouples image and text supervision so unpaired scans and reports can be used together, PMC-CLIP \cite{pmcclip} trains on figures and captions mined from the literature, and BiomedCLIP \cite{biomedclip} scales the recipe to fifteen million pairs. Two limits carry over to our setting: these models are validated with general-domain logic and never treat medical vision as a multi-modal expert domain, and ``biomedical'' spans many specialties with large knowledge gaps between them, dentistry being far more specialised than general medicine \cite{adams1999dentistry}. MedROV \cite{medrov} brings real-time open-vocabulary detection to nine medical modalities but relies on the general biomedical BiomedCLIP space, which is not adapted to the oral domain. We instead build the whole stack, vision tower, text tower, OralCLIP and OralDetect, from dental images and a dental corpus.

\textbf{Open-vocabulary detection.} Conventional detectors, two-stage \cite{fasterrcnn}, one-stage \cite{retinanet}, and DETR-style \cite{dino,rtdetr}, learn a fixed classifier over a fixed label set, so a new category needs a new head and new training data. Open-vocabulary detection replaces the classifier with text embeddings: the embedded class names are the prototypes, so the decision boundary depends on the image-text space they come from. OVR-CNN \cite{ovrcnn} learns region-word association from captions, ViLD \cite{vild} and RegionCLIP \cite{regionclip} distil a contrastive image-text space \cite{clip} into region features, Detic \cite{detic} adds image-level supervision, and OWL-ViT \cite{owlvit} attaches box heads to a contrastive backbone. GLIP \cite{glip} recasts detection as phrase grounding, and Grounding DINO \cite{groundingdino} fuses language into a DETR detector at several depths; such fusion performs well in-set, but its cross-attention is expensive and the whole vocabulary must be packed into one short sentence. YOLO-World \cite{yoloworld} and OmDet-Turbo \cite{omdetturbo} embed each class name independently and keep only a light text-guided fusion. Retrieval detectors drop the fusion and score regions directly against the class-name embeddings \cite{owlvit}, and WeDetect \cite{fu2026wedetect} shows that this design is stronger with better pretraining and much faster, with no length budget on the vocabulary. Naive retrieval detectors, however, suffer multi-modal semantic confusion in the oral domain; OralDetect keeps the fast retrieval design and calibrates the semantic shift with our modality-conditioned calibration module.

\textbf{Computer vision in the dental domain.} Dental image analysis has largely been developed one modality and task at a time, including tooth segmentation and FDI identification on panoramic radiographs \cite{toothsegpano}, pathology detection \cite{alphadent}, fenestration, dehiscence, and plaque detection on intraoral photographs \cite{fdtooth,plaquesam}, and tumour analysis in histopathology \cite{osccpatho}. These closed-vocabulary models require new task-specific labels and heads for unseen conditions and have a single-modality scope, while clinical diagnosis needs shared semantics across modalities. Recent work has moved toward multi-modal dental foundation models. MMOral \cite{mmoral} introduced a multimodal benchmark and foundation model, followed by OralGPT-Omni \cite{oralgptomni}, OralGPT-Plus \cite{oralgptplus}, and OralAgent \cite{oralagent}, extending coverage across modalities, tasks, tool use, and knowledge retrieval. However, these systems are primarily evaluated through visual question answering: they generate textual responses but struggle to localise clinical findings. Dentists often need to ask what a finding is and where it appears, regardless of the input modality. Yet no existing benchmark, detector, or dental foundation model jointly supports open-vocabulary detection across modalities. We address this gap with the OralDetect family, which unifies a dental vision-language space, a cross-modality detector, and a benchmark with a shared vocabulary and evaluation protocol.

\section{Data Curation and OralDetect-Bench}
\label{sec:data}

\begin{table}[!htbp]
\centering
\caption{Composition of the cleaned image-text corpus, in three tiers. (i) Diagnostic images from eight oral imaging modalities. (ii) General dental figures, e.g., textbook illustrations and schematics. (iii) General-domain images.}
\label{tab:corpus}
\footnotesize
\begin{adjustbox}{max width=\linewidth}
\begin{tabular}{lr}
\toprule
& \# image--text pairs \\
\midrule
\multicolumn{2}{l}{\emph{(i) Dental diagnostic imaging}} \\
\quad Intraoral Image                & 263,933 \\
\quad Panoramic X-ray                & 74,495 \\
\quad Periapical X-ray               & 63,862 \\
\quad Histopathology                 & 30,373 \\
\quad Cytology                       & 10,961 \\
\quad CT/CBCT                       & 9,516 \\
\quad Cephalometric X-ray            & 5,269 \\
\quad MRI                            & 8,542 \\
\quad \textit{subtotal}              & \textit{466,951} \\
\midrule
\multicolumn{2}{l}{\emph{(ii) Dental knowledge figures}} \\
\quad \textit{subtotal}              & \textit{251,888} \\
\midrule
{\emph{(iii) General-domain pairs (COCO)}} &  118,273 \\
\midrule
Full corpus                          & \textbf{837{,}112} \\
\bottomrule
\end{tabular}
\end{adjustbox}
\end{table}

We collect more than 1M images from public datasets and an in-house dataset to construct the OralDetect infrastructure; the resources are detailed in Appendix~\ref{sec:sources}. After removing duplicates and cleaning the captions, we curate more than 830K image-text pairs for our OralDetect suite (\cref{fig:main}~(c)). These data span 8 typical diagnostic imaging modalities, other dental pairs for dental-knowledge injection, and general pairs from COCO \cite{coco} to avoid overfitting and improve out-of-distribution ability (\cref{tab:corpus}). Based on the cleaned corpus, we construct OralDetect-Bench with two sub-benches, one for each half of the stack. The \textbf{retrieval bench} directly measures the image-text space of our pretrained OralCLIP and representative baselines. The \textbf{detection bench} assesses region-level localisation under a shared open vocabulary, both in-set and open-set (dental plaque and crack).

\begin{figure}[!htbp]
\centering
\includegraphics[width=\linewidth]{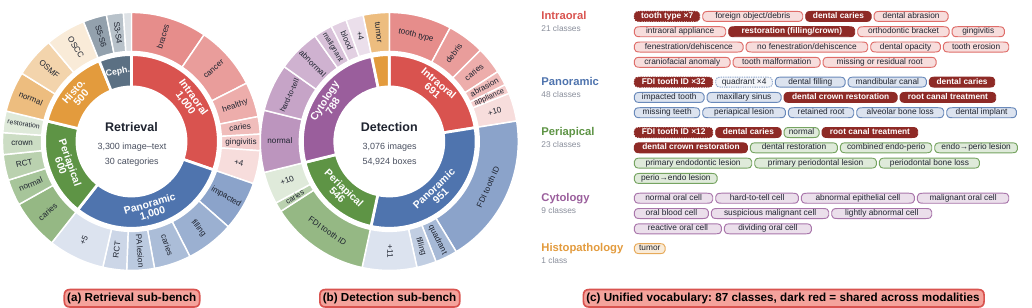}
\caption{\textbf{OralDetect-Bench at a glance.} \textbf{(a)} Retrieval and \textbf{(b)} detection sub-benches: inner ring = each modality's share of images, outer ring = its most frequent categories. \textbf{(c)} The unified 87-class vocabulary; dark chips are concepts shared across modalities, dashed chips group tooth-position labels (FDI IDs, quadrants, tooth types).}
\label{fig:bench}
\end{figure}

\subsection{Retrieval Bench}
\label{sec:bench_retrieval}

The retrieval bench (\cref{fig:bench}~(a)) holds 3,300 image-text pairs with 30 categories, drawn from 2--5 source datasets per modality, over five imaging modalities: intraoral photography (1,000), panoramic X-ray (1,000), periapical radiography (600), histopathology (500), and cephalometric X-ray (200). Every pair is excluded from the OralCLIP training pipeline. To keep the clinical information accurate and practical, captions are short, real clinical conditions taken from expert-annotated classification, detection, and segmentation tasks, rather than free descriptions. We state the modality in the text to add modality calibration implicitly, e.g., \textbf{Oral cavity intraoral photograph showing orthodontic braces}. All pairs also carry a category label, so this sub-bench supports three tasks following BiomedCLIP \cite{biomedclip}: (i) text-to-image retrieval, (ii) image-to-text retrieval, and (iii) zero-shot classification.

\subsection{Detection Bench}

\begin{table}[!htbp]
\centering
\caption{The detection sub-bench. The closed set is fixed per modality. The open-set data are excluded from training and validation and evaluated separately. \#cls: vocabulary size per modality.}
\label{tab:detbench}
\footnotesize
\setlength{\tabcolsep}{4pt}
\begin{adjustbox}{max width=\linewidth}
\begin{tabular}{lrrr}
\toprule
Modality & Images & Boxes & \#cls \\
\midrule
\multicolumn{4}{l}{\emph{In-set task}} \\
Intraoral Image  &   691 &  5{,}763 & 22 \\
Panoramic X-ray  &   951 & 41{,}852 & 48 \\
Periapical X-ray &   546 &  1{,}727 & 23 \\
Cytology         &   788 &  5{,}482 &  9 \\
Histopathology   &   100 &    100 &  1 \\
\midrule
\textbf{Closed-set total} & \textbf{3{,}076} & \textbf{54{,}924} & \textbf{87} \\
\midrule
\multicolumn{4}{l}{\emph{Open-set task}} \\
Intraoral (dental plaque)    &  48 & 816 & 1 \\
Intraoral (dental crack)     & 805 & 1{,}435 & 1 \\
\midrule
\textbf{Open-set total} & \textbf{853} & \textbf{2{,}251} & \textbf{2} \\
\bottomrule
\end{tabular}
\end{adjustbox}
\end{table}

The detection bench is assembled from 14 main public sources covering five dental imaging modalities (\cref{tab:detbench,fig:bench}). All datasets were annotated by experts; their reorganisation is illustrated in \cref{fig:main}~(c). We organise two tasks. \textbf{(i) Closed-set cross-modality detection} over the shared 87-class vocabulary: the source datasets are used for training, but the test samples are held out in advance. \textbf{(ii) Open-set detection} of \textit{dental plaque} and \textit{dental crack} on intraoral images, whose datasets are excluded from training and validation entirely, evaluating zero-shot open-vocabulary detection. As \cref{fig:bench}~(c) shows, the vocabulary is long-tailed and $15$ class names, including the most frequent findings, recur in several modalities, which is precisely the cross-modality ambiguity our calibration targets.

\subsection{Evaluation Protocol}
Both benchmarks are scored \emph{within} each modality, but no modality prior is ever given to the model at test time: the model must overcome semantic confusion among modalities without explicitly profiting from modality information. Retrieval reports Recall@$\{1,5,10\}$ in both directions and top-1 zero-shot classification accuracy. Detection reports mAP@[.5:.95] under COCO evaluation \cite{coco}, restricted per modality to that modality's images and ground-truth categories. We aggregate across modalities both by an unweighted macro average and an image-weighted micro average over the five modalities.

\section{OralDetect Family}
\label{sec:method}

\subsection{OralCLIP from Two Encoders}
\begin{figure}[!htbp]
\centering
\includegraphics[width=0.62\linewidth]{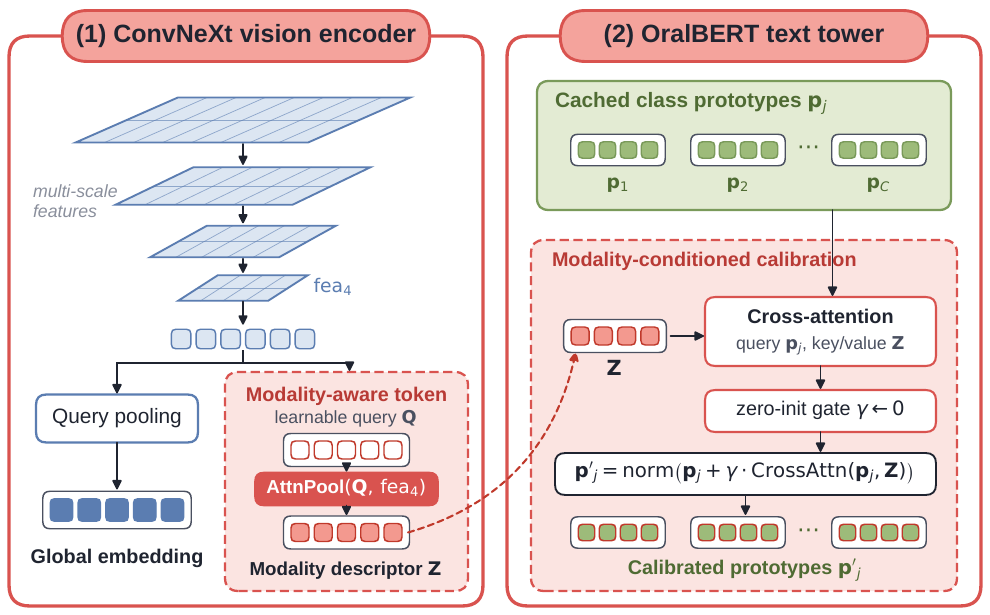}
\caption{OralCLIP architecture: \textbf{(1)} the ConvNeXt vision encoder with its modality-aware token, and \textbf{(2)} the OralBERT text tower with modality-conditioned prototype calibration.}
\label{fig:oralclip}
\end{figure}

To address the domain-adaptation gap and modality-related semantic confusion, we construct OralCLIP with an additional modality-calibration module and pretrain it on our curated image-text pairs instead of directly fine-tuning a general-domain CLIP model (\cref{fig:oralclip}).

\textbf{Multi-scale vision encoder.} Oral findings vary widely in the fraction of the image they occupy: a periodontal pattern spans an entire arch, a carious lesion a few millimetres of enamel, a malignant nucleus a single cell in a cytology field. A detector for oral imaging must therefore read evidence across a wide range of scales. A plain vision transformer commits to one patch granularity and must learn scale invariance implicitly. We instead use ConvNeXt-Base~\cite{convnext}, whose hierarchical pyramid emits multi-scale feature maps by construction, and initialise it from the WeDetect open-vocabulary pretrained weights~\cite{fu2026wedetect}, so our OralConv begins from a representation already tuned for localisation.

\textbf{OralBERT.} The text tower carries the class semantics, and current biomedical encoders know little dentistry. We build OralBERT by continuing masked-language-model pretraining of PubMedBERT~\cite{biomedbert} on a corpus of $368$ dental textbooks ($\approx\!47.7$M tokens under the PubMedBERT tokenizer), injecting the vocabulary and terms used by the downstream detector. Its evaluation is reported in Appendix~\ref{sec:oralbert}.

\textbf{Contrastive alignment with a modality-aware head.} OralCLIP aligns the two towers with an InfoNCE objective and a learnable temperature into a shared, $\ell_2$-normalised embedding space, over our corpus spanning 8 oral imaging modalities and an ``other'' category (\cref{tab:corpus}). A plain contrastive loss over a pooled multi-modality corpus, however, cannot adapt to the semantic bias each modality injects. The same caption is paired with visually incompatible images across modalities, and the objective has no mechanism to separate what a finding means from how it looks in a given modality. We therefore add a modality-aware auxiliary head. Because ConvNeXt is a convolutional network with no token stream, we realise the special modality token as a learnable query that attention-pools the deepest feature map into a per-image modality descriptor. A class-weighted cross-entropy on the resulting modality logits (weight $0.15$) supervises it, pushing the vision features to encode modality explicitly rather than leak it into the alignment.

\subsection{Open-Vocabulary OralDetect}
\textbf{Fusion entangles image and text.} Grounding DINO~\cite{groundingdino} and related fusion detectors align vision and language by passing the image tokens $V$ and the caption tokens $T$ through stacked cross-attention layers, so the region features that feed the box head are conditioned on the caption:
\begin{equation}
(\tilde{V},\,\tilde{T}) = \mathrm{Fuse}(V,\,T), \qquad
s_{ij} = \langle\, \tilde{V}_i,\ \tilde{T}_j \,\rangle .
\label{eq:fusion}
\end{equation}
Both $\tilde{V}$ and $\tilde{T}$ depend on the \emph{joint} input $(V, T)$, so the text side cannot be embedded once and reused: changing the vocabulary $T\!\to\!T'$ re-runs the entire fused backbone at $O(|V|\,|T|\,L)$ per image-vocabulary pair, and neither tower yields a reusable standalone representation.

\textbf{Detection as retrieval.} Inspired by WeDetect~\cite{fu2026wedetect}, we turn detection into retrieval. Our vision tower $\Phi$ (OralConv) produces region features, and the pretrained text tower $\Psi$ (OralBERT) embeds each class name into a prototype independently, scored by cosine similarity:
\begin{equation}
\mathbf{r}_i = \Phi(I)_i, \quad \mathbf{p}_j = \Psi(c_j), \quad
s_{ij} = \frac{\langle\, \mathbf{r}_i,\ \mathbf{p}_j \,\rangle}{\|\mathbf{r}_i\|\,\|\mathbf{p}_j\|},
\label{eq:retrieval}
\end{equation}
with $\Phi$ independent of the class names and $\Psi$ independent of the image. The $C$ prototypes $\{\mathbf{p}_j\}$ are computed once and cached for the entire test set, so adding or removing a class only re-embeds a single string and leaves the image forward untouched. The cost drops from fusion's $O(|V|\,|T|\,L)$ to $O(|V|)+O(C)$, and the head is agnostic to the vocabulary size. The towers stay modular, so OralCLIP drops in unchanged. The class-name embeddings \emph{are} the classifier, so any change to the text tower moves the decision boundary, which is why a dental-knowledgeable OralBERT matters.

\textbf{Modality-conditioned calibration.} A retrieval design, however, ignores the modality gap. A single prototype $\mathbf{p}_j$ per name is shared across modalities, yet one name denotes different visual evidence in each, e.g., \emph{dental caries} is a cavitated enamel surface in an intraoral photograph and a radiolucent shadow on a periapical radiograph. The prototype that minimises the contrastive loss is pulled toward an average over modalities and matches none well. We make each prototype a light function of the image being scored. We attention-pool the deepest vision map with $M$ learnable queries $Q$ into a per-image modality descriptor
\begin{equation}
\mathbf{Z} = \mathrm{AttnPool}\big(Q,\ \Phi_{\mathrm{deep}}(I)\big) \in \mathbb{R}^{M\times d},
\label{eq:modtoken}
\end{equation}
so modality is inferred without any explicit modality input, adapted from the special token used in OralCLIP pretraining. The prototypes then cross-attend to $\mathbf{Z}$ (query $=$ prototype, key $=$ value $=\mathbf{Z}$) through a zero-initialised gate:
\begin{equation}
\mathbf{p}'_j = \mathrm{normalise}\big(\mathbf{p}_j + \gamma\,\mathrm{CrossAttn}(\mathbf{p}_j,\,\mathbf{Z})\big),
\qquad \gamma \!\leftarrow\! 0,
\label{eq:calib}
\end{equation}
and the score is $s_{ij} = \langle \bar{\mathbf{r}}_i,\ \bar{\mathbf{p}}'_j\rangle$ on the $\ell_2$-normalised vectors. At step $0$ the module is exactly the identity, so the pretrained text tower is untouched and the calibration only earns its way in. Crucially, the expensive $\Psi$ forward is still cached and only the tiny $C\times M$ cross-attention re-runs per image, so we keep the $O(|V|)+O(C)$ efficiency of retrieval while the prototype becomes modality-conditioned, i.e., $\mathbf{p}'_j = \mathbf{p}'_j(I)$ for each input image $I$.

\section{Experiments}
\label{sec:exp}

\begin{table}[!htbp]
\centering
\caption{Results on the detection bench (3,076 images, 87 classes). $Avg_{mac}$ is the unweighted mean of the five per-modality mAPs; $Avg_{mic}$ is the image-weighted mean over the five modalities.}
\label{tab:main}
\footnotesize
\setlength{\tabcolsep}{4pt}
\begin{adjustbox}{max width=\linewidth}
\begin{tabular}{llcccccccc}
\toprule
& & & \multicolumn{5}{c}{mAP@[.5:.95] per modality} & & \\
\cmidrule(lr){4-8}
Method & Vision / text tower & Vocab. & Intra. & Pano. & Peri. & Cyto. & Histo. & $Avg_{mac}$ & $Avg_{mic}$ \\
\midrule
Faster R-CNN \cite{fasterrcnn} & R50-FPN / --- & closed & 0.543 & 0.590 & 0.472 & 0.247 & 0.608 & 0.492 & 0.471 \\
RetinaNet \cite{retinanet}     & R50-FPN / --- & closed & 0.484 & 0.503 & 0.362 & 0.196 & 0.678 & 0.445 & 0.401 \\
DINO \cite{dino}               & R50 / ---     & closed & 0.535 & 0.796 & 0.488 & 0.237 & 0.699 & 0.551 & 0.537 \\
RT-DETR \cite{rtdetr}          & R50 / ---     & closed & 0.522 & 0.739 & 0.472 & 0.242 & 0.653 & 0.526 & 0.513 \\
\midrule
OWL-ViT \cite{owlvit} & ViT-B/32 / CLIP     & open & 0.444 & 0.311 & 0.484 & 0.134 & 0.756 & 0.426 & 0.341 \\
Detic \cite{detic}                & R50 / CLIP          & open & 0.490 & 0.612 & 0.460 & 0.226 & 0.710 & 0.500 & 0.462 \\
GLIP \cite{glip}     & Swin-T / BERT       & open & 0.532 & 0.794 & 0.495 & 0.270 & 0.011 & 0.420 & 0.522 \\
Grounding DINO \cite{groundingdino} & Swin-T / BERT     & open & 0.247 & 0.228 & 0.392 & 0.153 & 0.614 & 0.327 & 0.255 \\
YOLO-World \cite{yoloworld}       & YOLOv8-L / CLIP     & open & 0.501 & 0.671 & 0.424 & 0.228 & 0.638 & 0.492 & 0.474 \\
OmDet-Turbo \cite{omdetturbo} & Swin-T / CLIP & open & 0.494 & 0.405 & 0.489 & 0.241 & \textbf{0.765} & 0.479 & 0.409 \\
OV-DINO \cite{ovdino}             & Swin-T / BERT       & open & 0.497 & 0.668 & 0.472 & 0.301 & 0.258 & 0.439 & 0.488 \\
MedROV \cite{medrov}              & YOLOv8-L / BiomedCLIP & open & 0.358 & 0.550 & 0.277 & 0.114 & 0.084 & 0.276 & 0.332 \\
\rowcolor{oursred!10}
\textbf{OralDetect (ours)} & OralCLIP-B / OralBERT & open & \textbf{0.603} & \textbf{0.805} & \textbf{0.524} & \textbf{0.331} & 0.598 & \textbf{0.572} & \textbf{0.582} \\
\bottomrule
\end{tabular}
\end{adjustbox}
\end{table}

\subsection{Implementation}
Our detector follows the WeDetect design~\cite{fu2026wedetect}: a ConvNeXt-Base vision tower~\cite{convnext}, a BERT-family text tower~\cite{bert}, a CSP-RepBiFPAN neck~\cite{yolov6}, and a YOLO-World-style region-text alignment head~\cite{yoloworld}, with inherited hyperparameters, plus our lightweight modality-calibration module. Targets are assigned by a task-aligned assigner~\cite{tood}, and the loss sums a region-text binary cross-entropy ($0.5$), CIoU~\cite{ciou} ($7.5$) and distribution focal loss~\cite{gfl} ($0.375$). We optimise with AdamW (weight decay $0.05$, gradient clip $\ell_2\!=\!10$) under a linear-warmup cosine schedule~\cite{sgdr}, using mosaic~\cite{yolov4}, mixup~\cite{mixup}, random affine, HSV jitter, and horizontal flip. We initialise from our pretrained \textsc{OralCLIP} vision tower and \textsc{OralBERT} checkpoint, build on MMDetection and MMYOLO~\cite{mmdetection,mmyolo}, and evaluate under the COCO protocol~\cite{coco} on the held-out test images.

\subsection{Baselines}
We compare against two families, all evaluated under the same per-modality COCO protocol and trained on the same OralDetect split as our model. \textbf{Closed-set detectors}, Faster R-CNN~\cite{fasterrcnn}, RetinaNet~\cite{retinanet}, DINO~\cite{dino} and RT-DETR~\cite{rtdetr}, learn a fixed $87$-class classifier. \textbf{General open-vocabulary detectors}, OWL-ViT~\cite{owlvit}, Detic~\cite{detic}, GLIP~\cite{glip}, Grounding DINO~\cite{groundingdino}, YOLO-World~\cite{yoloworld}, OmDet-Turbo~\cite{omdetturbo} and OV-DINO~\cite{ovdino}, receive the $87$ class names as the text prompt. MedROV~\cite{medrov} is an open-vocabulary detector trained specifically for the medical domain on the BiomedCLIP space.

\begin{figure}[!htbp]
\centering
\includegraphics[width=\linewidth]{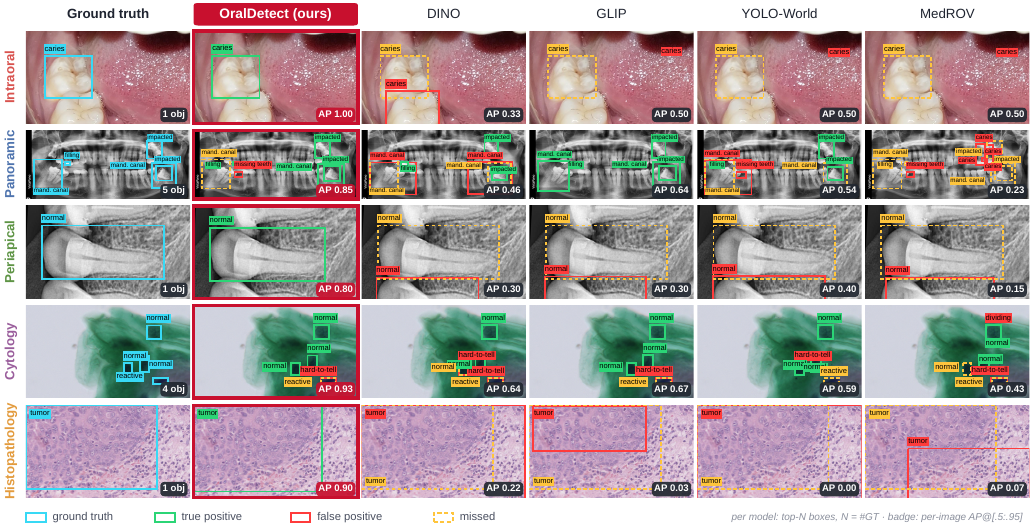}
\caption{\textbf{Qualitative comparison}, one test image per modality. Each model shows its top-$N$ detections ($N$ = number of ground-truth boxes): \textcolor[HTML]{1FA85A}{\textbf{green}} = correct (IoU${\ge}0.5$), \textcolor[HTML]{E03131}{\textbf{red}} = false positive, \textcolor[HTML]{D49A00}{\textbf{amber dashed}} = missed; each box carries its (predicted or missed) class; badges give per-image AP@[.5:.95]. Panoramic shows findings only. Images were chosen among those where OralDetect scores highest.}
\label{fig:qual}
\end{figure}

\begin{table}[!htbp]
\centering
\caption{Ablations: removing the modality calibration module, and replacing our pretrained foundation models with a generic initialisation. Bold marks the best in each column of the closed-set bench.}
\label{tab:ablation}
\footnotesize
\setlength{\tabcolsep}{5pt}
\begin{adjustbox}{max width=\linewidth}
\begin{tabular}{lccccccc}
\toprule
Setting & Intraoral & Panoramic & Periapical & Cytology & Histopathology & $Avg_{mac}$ & $Avg_{mic}$  \\
\midrule
\rowcolor{oursred!10}
Our OralDetect & \textbf{0.603} & \textbf{0.805} & \textbf{0.524} & \textbf{0.331} & 0.598 & \textbf{0.572} & \textbf{0.582} \\
\quad $-$ modality calibration   & 0.560 & 0.761 & 0.506 & 0.251 & \textbf{0.742} & 0.564 & 0.539 \\
\quad $-$ pretrained foundations    & 0.141 & 0.320 & 0.251 & 0.024 & 0.581 & 0.263 & 0.200 \\
\bottomrule
\end{tabular}
\end{adjustbox}
\end{table}

\subsection{Main Results}
\Cref{tab:main} reports per-modality mAP@[.5:.95] on the closed-set detection bench. OralDetect attains the best macro ($0.572$) and micro ($0.582$) average against every open-vocabulary baseline and the closed-set specialists, whose strongest is DINO at $0.551$ macro. Our open-vocabulary model thus outperforms category-specific detectors while retaining free-form vocabulary detection. It ranks first on four of the five modalities: intraoral ($0.603$), panoramic ($0.805$), periapical ($0.524$), and cytology ($0.331$).
Fusion detectors are the most modality-fragile: GLIP is strong on panoramic ($0.794$) but collapses on histopathology ($0.011$), and Grounding DINO has the lowest micro mAP ($0.255$), their deep image-text fusion overfitting the dominant modality and failing on the long-tail ones. The detectors that embed class names independently, ours, YOLO-World, and OmDet-Turbo, are far more even across modalities, supporting the two-tower design, while MedROV has the lowest macro mAP ($0.276$), showing that a general biomedical space transfers poorly to dentistry. Histopathology is the one exception to our lead: OmDet-Turbo ($0.765$) does best, but our $0.598$ stays stable where other open-vocabulary models degenerate (GLIP $0.011$, MedROV $0.084$, OV-DINO $0.258$).
\Cref{fig:qual} makes these differences visible. On the periapical film, all four baselines name the right class but misplace the box, while OralDetect localises it; on the cytology field, they confuse visually similar cell types, the fine-grained ambiguity our domain pretraining and calibration are designed to target.

\subsection{Ablation Study}
\Cref{tab:ablation} isolates the two main contributions. Removing the pretrained OralCLIP/OralBERT foundations and training the same architecture from a generic initialisation is by far the larger effect: macro mAP falls from $0.572$ to $0.263$ and micro from $0.582$ to $0.200$, showing that a well-curated dental image-text corpus and a suite of dedicated foundation models are essential in this domain.
The modality-conditioned calibration improves four of the five modalities ($+0.008$ macro, $+0.043$ micro). Its benefit concentrates where one name maps to different visual evidence across modalities: cytology ($0.251$$\!\to\!0.331$), panoramic ($0.761$$\!\to\!0.805$), and intraoral ($0.560\!\to\!0.603$). On the long-tail histopathology set, which carries no such cross-modality ambiguity, it is counterproductive ($0.742\!\to\!0.598$); we expect a larger and more balanced dataset to reveal the recipe's full potential, and the net effect remains positive.

\begin{table}[!htbp]
\centering
\caption{Open-set detection (Precision, F1); best in \textbf{bold}. Dental crack: all output boxes kept.}
\label{tab:plaque_prf}
\footnotesize
\setlength{\tabcolsep}{4pt}
\begin{adjustbox}{max width=\linewidth}
\begin{tabular}{lcccc}
\toprule
& \multicolumn{2}{c}{Dental plaque} & \multicolumn{2}{c}{Dental crack} \\
\cmidrule(lr){2-3}\cmidrule(lr){4-5}
Model & Precision & F1 & Precision & F1 \\
\midrule
OWL-ViT              & 0.224 & 0.172 & 0.018 & 0.035 \\
Detic                & 0.212 & 0.206 & 0.021 & 0.029 \\
GLIP                 & 0.189 & 0.184 & 0.017 & 0.032 \\
Grounding DINO       & 0.258 & 0.036 & \textbf{0.361} & 0.288 \\
YOLO-World           & 0.207 & 0.155 & 0.069 & 0.009 \\
OmDet-Turbo          & 0.174 & 0.179 & 0.018 & 0.035 \\
OV-DINO              & 0.204 & 0.240 & 0.006 & 0.012 \\
MedROV               & 0.159 & 0.144 & 0.000 & 0.000 \\
\rowcolor{oursred!10}
\textbf{OralDetect (ours)} & \textbf{0.286} & \textbf{0.282} & 0.212 & \textbf{0.333} \\
\bottomrule
\end{tabular}
\end{adjustbox}
\end{table}

\subsection{Open-Set Generalisation}
To probe transfer beyond the training distribution, we evaluate on the held-out dental-plaque dataset \cite{roboflowplaque} (\cref{tab:plaque_prf}), never seen in training or validation. This mimics clinical use, where a user inputs an ordinary intraoral image and asks whether a condition appears. We keep boxes with confidence $\ge0.3$ and count a detection as correct at IoU$\ge0.5$. Diffuse plaque on intraoral photographs is hard but common in daily practice. OralDetect leads on both precision ($0.286$) and F1 ($0.282$), ahead of the next-best OV-DINO ($0.240$ F1), while Grounding DINO barely fires ($0.036$ F1). As on the closed-set bench, OralDetect ranks first, indicating that the dental image-text space, not memorised categories, drives the gain.
On the 805 box-annotated photos (1,435 boxes) of a public dental-crack set \cite{dentalcrack}, OralDetect has the best F1 ($0.333$, recall $0.782$, IoU$\ge0.1$); Grounding DINO is more precise ($0.361$) but finds only $24\%$ of cracks, and OralDetect still leads at IoU$\ge0.3$ ($0.085$ vs.\ $0.065$ F1).

\subsection{Retrieval Results}
We evaluate our foundation models on the retrieval sub-bench (\cref{tab:retrieval_results}), assessing the OralCLIP image-text space against typical CLIP-style models: CLIP \cite{clip}, trained on general image-text pairs, and MedCLIP \cite{medclip}, PubMedCLIP \cite{pubmedclip}, and BiomedCLIP \cite{biomedclip}, trained on medical pairs. OralCLIP tops every column and panel, with mean R@1 of $0.712$ image-to-text, $0.791$ text-to-image and $0.763$ top-1 zero-shot accuracy, against $0.633/0.703/0.675$ for the strongest baseline, BiomedCLIP fine-tuned on the identical corpus (BiomedCLIP-ft); the gap to it isolates the contribution of our architecture and modality-aware training. OralCLIP beats the general and biomedical CLIPs by a large margin (e.g., CLIP $0.311$ image-to-text mean). Panoramic is the hardest modality, since its multi-label captions depress every method, yet OralCLIP still leads ($0.544$).

\begin{table}[!htbp]
\centering
\caption{Results on the retrieval sub-bench: R@1 for cross-modal retrieval in both directions and top-1 accuracy for zero-shot classification, per modality. BiomedCLIP-ft and OralCLIP are both trained on the OralCLIP corpus. Best in each column and panel in bold. Panoramic is multi-label ($10$ categories occur in $281$ combinations), so it has no zero-shot score and that mean covers four modalities.}
\label{tab:retrieval_results}
\footnotesize
\setlength{\tabcolsep}{6pt}
\begin{adjustbox}{max width=\linewidth}
\begin{tabular}{lcccccc}
\toprule
Model & Intraoral & Panoramic & Periapical & Histopathology & Cephalometric & Mean \\
\midrule
\multicolumn{7}{l}{\emph{Image$\rightarrow$Text retrieval (R@1)}} \\
CLIP \cite{clip}              & 0.450 & 0.084 & 0.352 & 0.336 & 0.335 & 0.311 \\
BiomedCLIP \cite{biomedclip}  & 0.451 & 0.069 & 0.352 & 0.514 & 0.405 & 0.358 \\
PubMedCLIP \cite{pubmedclip}  & 0.556 & 0.133 & 0.332 & 0.240 & 0.295 & 0.311 \\
MedCLIP \cite{medclip}        & 0.137 & 0.130 & 0.185 & 0.292 & 0.310 & 0.211 \\
BiomedCLIP-ft                 & 0.833 & 0.443 & 0.665 & 0.944 & 0.280 & 0.633 \\
\rowcolor{oursred!10}
\textbf{OralCLIP (ours)}      & \textbf{0.887} & \textbf{0.544} & \textbf{0.767} & \textbf{0.952} & \textbf{0.410} & \textbf{0.712} \\
\midrule
\multicolumn{7}{l}{\emph{Text$\rightarrow$Image retrieval (R@1)}} \\
CLIP \cite{clip}              & 0.262 & 0.161 & 0.446 & 0.631 & 0.286 & 0.357 \\
BiomedCLIP \cite{biomedclip}  & 0.386 & 0.195 & 0.486 & 0.930 & 0.143 & 0.428 \\
PubMedCLIP \cite{pubmedclip}  & 0.163 & 0.090 & 0.385 & 0.331 & 0.357 & 0.265 \\
MedCLIP \cite{medclip}        & 0.099 & 0.111 & 0.115 & 0.127 & 0.357 & 0.162 \\
BiomedCLIP-ft                 & 0.896 & 0.447 & 0.818 & \textbf{1.000} & 0.357 & 0.703 \\
\rowcolor{oursred!10}
\textbf{OralCLIP (ours)}      & \textbf{0.985} & \textbf{0.573} & \textbf{0.899} & \textbf{1.000} & \textbf{0.500} & \textbf{0.791} \\
\midrule
\multicolumn{7}{l}{\emph{Zero-shot classification (top-1 Accuracy)}} \\
CLIP \cite{clip}              & 0.331 & --- & 0.247 & 0.410 & 0.255 & 0.311 \\
BiomedCLIP \cite{biomedclip}  & 0.374 & --- & 0.340 & 0.492 & 0.235 & 0.360 \\
PubMedCLIP \cite{pubmedclip}  & 0.539 & --- & 0.218 & 0.302 & 0.240 & 0.325 \\
MedCLIP \cite{medclip}        & 0.117 & --- & 0.138 & 0.364 & 0.320 & 0.235 \\
BiomedCLIP-ft                 & 0.837 & --- & 0.597 & 0.938 & 0.330 & 0.675 \\
\rowcolor{oursred!10}
\textbf{OralCLIP (ours)}      & \textbf{0.857} & --- & \textbf{0.698} & \textbf{0.946} & \textbf{0.550} & \textbf{0.763} \\
\bottomrule
\end{tabular}
\end{adjustbox}
\end{table}

\section{Conclusion}
We presented the OralDetect family: dental vision and text towers aligned by OralCLIP, a retrieval-based open-vocabulary detector, and OralDetect-Bench for cross-modality retrieval and detection. Across five modalities, OralDetect surpasses closed-set and open-vocabulary baselines, OralCLIP leads every retrieval panel, and the detector generalises to two held-out open-set conditions. All data, benchmarks, and models will be released. Limitations are the small, single-class histopathology set and the small, intraoral-only open sets.

\bibliographystyle{plainnat}
\bibliography{refs}
\ogbeginappendix

\section{Source Datasets}
\label{sec:sources}

The detection library is assembled from 14 main public sources across the five modalities, listed as 15 entries because Rashid et al.\ contributes both colour and periapical images
(Table~\ref{tab:sources}). The counts in both tables are those we use after filtering and box conversion, so they can differ from the sizes the sources report.

\begin{table}[!htbp]
\centering
\caption{The 14 sources (15 entries) that form the detection library. Counts are those we use after filtering and box conversion and can differ from the sizes the sources report; AlphaDent, for example, is described by its authors as $1{,}320$ photographs, while the release we use holds $3{,}342$ annotated images.}
\label{tab:sources}
\footnotesize
\setlength{\tabcolsep}{5pt}
\begin{adjustbox}{max width=\linewidth}
\begin{tabular}{lllrrl}
\toprule
Modality & Source dataset & Access & Images & Boxes & Reference \\
\midrule
\multirow{7}{*}{Intraoral}
 & AlphaDent            & preprint      & 3{,}342  & 16{,}097 & \cite{alphadent} \\
 & OMNI                 & GitHub        & 3{,}594  & 23{,}031 & \cite{omni2025} \\
 & IO150K               & CVPR'24       & 856      & 17{,}714 & \cite{io150k} \\
 & Gingivitis           & Mendeley (CC BY) & 538 & 2{,}270 & \cite{gingivitis2024} \\
 & FDTooth              & PhysioNet     & 241      & 2{,}892  & \cite{fdtooth} \\
 & Rashid et al.\ 2022 (colour) & PeerJ CS & 89   & 414      & \cite{rashid2022} \\
 & Identify-Teeth       & Roboflow (CC BY) & 83    & 2{,}114  & \cite{identifyteeth} \\
\midrule
Panoramic
 & MM-Oral-OPG           & HuggingFace  & 18{,}978 & 835{,}704 & \cite{mmoral} \\
\midrule
\multirow{4}{*}{Periapical}
 & Azleen et al.\ 2024   & Roboflow  & 3{,}736  & 30{,}477 & \cite{tooth-numbering_dataset} \\
 & Periapical-6diseases & Kaggle      & 1{,}831  & 2{,}189  & \cite{periapical6} \\
 & Fatima et al.\ 2023  & on request    & 511      & 804      & \cite{fatima2022} \\
 & Rashid et al.\ 2022  & PeerJ CS      & 120      & 125      & \cite{rashid2022} \\
\midrule
\multirow{2}{*}{Cytology}
 & UFSC OCPap             & Mendeley (CC BY-NC) & 6{,}779 & 33{,}152 & \cite{ufscocpap} \\
 & Oral cytology (Kaggle) & Kaggle      & 739      & 20{,}020 & \cite{kagglecyto} \\
\midrule
Histology
 & OCDC                 & Mendeley (CC BY) & 451   & 451      & \cite{ocdc} \\
\midrule
\multicolumn{3}{l}{\textbf{Total (15 entries, 14 sources)}} & \textbf{41{,}888} & \textbf{987{,}454} & \\
\bottomrule
\end{tabular}
\end{adjustbox}
\end{table}

\section{OralDetect Data Splits}
\label{sec:splits}

Table~\ref{tab:splits} gives the full split. The main text reports the test bench only, every other cell is reported here. 

\begin{table}[!htbp]
\centering
\caption{Full train/validation/test split of OralDetect, images and boxes per modality. The test
column is the union of the five fixed per-modality benches and is the benchmark reported in the main
text.}
\label{tab:splits}
\footnotesize
\setlength{\tabcolsep}{6pt}
\begin{adjustbox}{max width=\linewidth}
\begin{tabular}{lrrrrrrr}
\toprule
& & \multicolumn{2}{c}{Train} & \multicolumn{2}{c}{Validation} & \multicolumn{2}{c}{Test} \\
\cmidrule(lr){3-4}\cmidrule(lr){5-6}\cmidrule(lr){7-8}
Modality & \#cls & Images & Boxes & Images & Boxes & Images & Boxes \\
\midrule
Intraoral   & 22 & 6{,}874  & 50{,}989  &   969   & 7{,}034  & 691   & 5{,}763  \\
Panoramic   & 48 & 15{,}718 & 692{,}325 & 2{,}260 & 99{,}398 & 951   & 41{,}852 \\
Periapical  & 23 & 5{,}038  & 28{,}610  &   560   & 3{,}178  & 546   & 1{,}727  \\
Cytology    &  9 & 5{,}870  & 41{,}489  &   848   & 6{,}134  & 788   & 5{,}482  \\
Histology   &  1 & 316      & 316       &    35   & 35       & 100   & 100      \\
\midrule
Unified & \textbf{87} & \textbf{33{,}816} & \textbf{813{,}729} & \textbf{4{,}672} & \textbf{115{,}779} & \textbf{3{,}076} & \textbf{54{,}924} \\
\bottomrule
\end{tabular}
\end{adjustbox}
\end{table}

\section{OralBERT}
\label{sec:oralbert}

OralBERT is obtained by continuing masked-language-model pretraining of a biomedical encoder
\cite{biomedbert} on a corpus of $368$ dental and oral textbooks with $337{,}777$ chunks, approximately $250$M characters, $\approx\!47.7$M tokens under the source tokenizer. Training runs for 3 epochs at learning rate $5\times10^{-5}$ with an effective batch of 128, a cosine schedule and fp16, for $7{,}761$ optimiser steps. Two percent of the corpus is held out for validation and the remaining $98\%$ is used for training, under the same seed.

Table~\ref{tab:oralbert-ppl} reports masked-language-model perplexity and Table~\ref{tab:oralbert-cloze} a dental cloze test of $5{,}000$ items built from $495$ dental terms that are single tokens under the encoder's vocabulary. Domain adaptation lowers perplexity on the
oral corpus by $28\%$ and raises cloze top-1 accuracy by $12.7$ points. We report the cost as well as the gain, while perplexity on general PubMed abstracts rises slightly.

\begin{table}[!htbp]
\centering
\caption{Masked-language-model perplexity, $3{,}000$ held-out oral samples and $1{,}000$ PubMed
abstracts. Lower is better.}
\label{tab:oralbert-ppl}
\footnotesize
\setlength{\tabcolsep}{6pt}
\begin{adjustbox}{max width=\linewidth}
\begin{tabular}{lrr}
\toprule
Evaluation corpus & Base encoder & OralBERT \\
\midrule
Oral corpus (held-out 2\%) $\downarrow$ & 3.74 & \textbf{2.70} \\
PubMed abstracts $\downarrow$           & \textbf{2.35} & 2.52 \\
\bottomrule
\end{tabular}
\end{adjustbox}
\end{table}

\begin{table}[!htbp]
\centering
\caption{Oral cloze test: $5{,}000$ items over $495$ single-token dental terms, scored by top-1 and top-5 accuracy.}
\label{tab:oralbert-cloze}
\footnotesize
\setlength{\tabcolsep}{6pt}
\begin{adjustbox}{max width=\linewidth}
\begin{tabular}{lrr}
\toprule
Metric & Base encoder & OralBERT \\
\midrule
Top-1 accuracy $\uparrow$ & 0.3614 & \textbf{0.4882} \\
Top-5 accuracy $\uparrow$ & 0.6136 & \textbf{0.7622} \\
\bottomrule
\end{tabular}
\end{adjustbox}
\end{table}

\section{Qualitative Results}
\label{sec:qualitative}

\begin{figure}[!htbp]
\centering
\includegraphics[width=\linewidth]{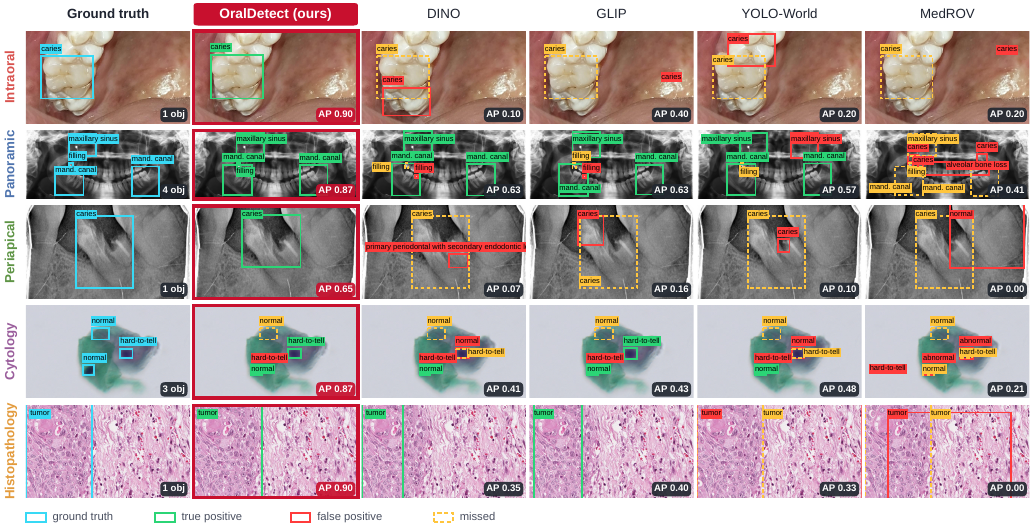}
\caption{Cases where OralDetect leads the strongest baseline by the largest margin (one test image per modality, excluding the main-text examples). Cyan = ground truth, green = true positive, red = false positive, amber dashed = missed; each box carries its (predicted or missed) class. Badges give per-image AP@[.5:.95].}
\label{fig:qualwin}
\end{figure}

\begin{figure}[!htbp]
\centering
\includegraphics[width=\linewidth]{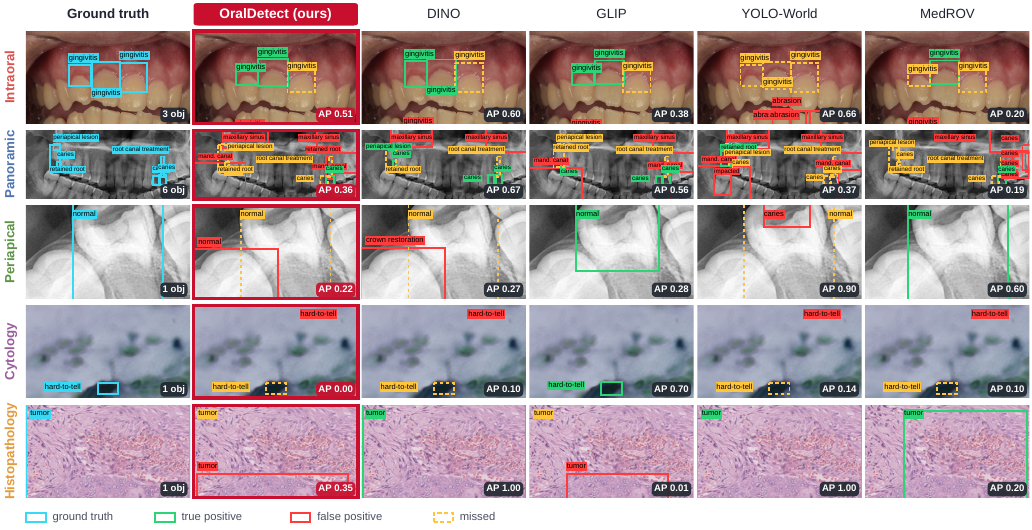}
\caption{Representative cases where OralDetect trails the strongest baseline. Colours and badges follow the same convention as in \cref{fig:qualwin}.}
\label{fig:qualloss}
\end{figure}

Figures~\ref{fig:qualwin} and~\ref{fig:qualloss} extend the qualitative comparison of the main text with the same layout and baselines. For each modality, Figure~\ref{fig:qualwin} shows the test image (other than those in the main text) on which OralDetect leads the strongest baseline by the largest per-image AP margin, and Figure~\ref{fig:qualloss} representative images on which it trails the strongest baseline. Intraoral: our single box covers the carious tooth, while the baselines miss it or place it elsewhere. Panoramic: all four findings, including the filling and both mandibular canals, are recovered, whereas the baselines add false fillings and caries. Periapical: the carious lesion is localised as one box, while the baselines mislocalise it or confuse it with other conditions. The failure cases include a missed \emph{gingivitis} region, a mislocalised \emph{normal} periapical region, a missed \emph{hard-to-tell} cell in a blurred cytology image, and a mislocalised tumour region.

\section{Open-Vocabulary Qualitative Results}
\label{sec:openset_qual}

\begin{figure}[!htbp]
\centering
\includegraphics[width=\linewidth]{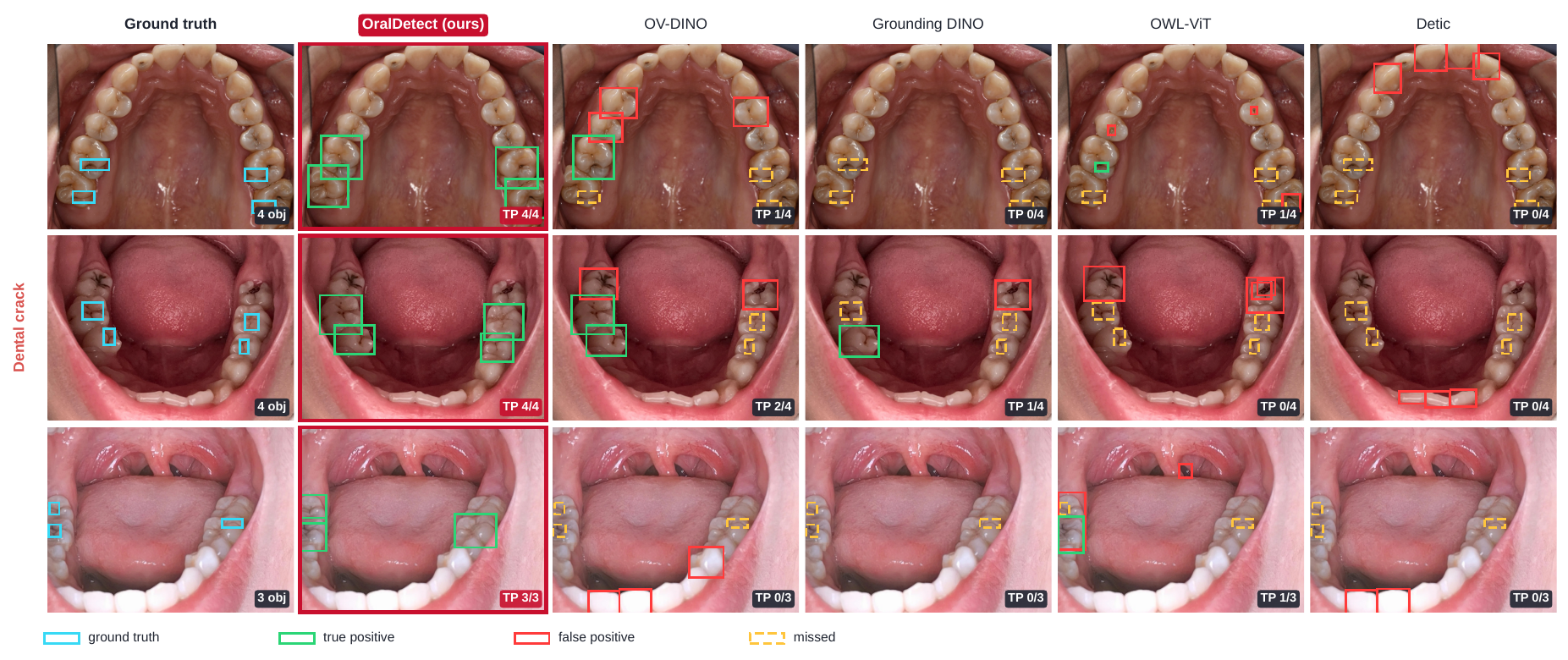}
\caption{Open-vocabulary detection on the held-out dental-crack set; images where OralDetect leads by the largest margin. Cyan = ground truth, green = true positive, red = false positive, amber dashed = missed.}
\label{fig:openset}
\end{figure}

\cref{fig:openset} compares OralDetect with the open-vocabulary baselines on the 805 box-annotated images of the held-out dental-crack set \cite{dentalcrack}, whose box annotations we use as released. OralDetect finds every annotated crack in these images, whereas the baselines miss most of them or fire on unrelated teeth.

\end{document}